\documentclass{article}

\usepackage{microtype}
\usepackage{graphicx}
\usepackage{multirow}
\usepackage{subfigure}
\usepackage{booktabs} % for professional tables
\usepackage{tikz}
\usetikzlibrary{positioning}

\usepackage{hyperref}

\usepackage[accepted]{icml2025}

\usepackage{amsmath}
\usepackage{amssymb}
\usepackage{mathtools}
\usepackage{amsthm}

\usepackage[capitalize,noabbrev]{cleveref}

\DeclareMathOperator{\normal}{\mathcal{N}}
\DeclareMathOperator{\expect}{\mathbb{E}}
\DeclareMathOperator{\mse}{MSE}
\DeclareMathOperator{\amse}{AMSE}
\DeclareMathOperator{\harm}{\mathsf{Y}}
\DeclareMathOperator{\psd}{PSD}
\DeclareMathOperator{\coh}{Coh}
\DeclareMathOperator{\real}{\mathfrak{R}}
\DeclareMathOperator{\crps}{CRPS}
\DeclareMathOperator{\erms}{eRMSE}
\DeclareMathOperator{\uberms}{ub\_eRMSE}
\DeclareMathOperator{\ser}{SER}
\DeclareMathOperator{\dkl}{D_\mathrm{KL}}
\allowdisplaybreaks % Allow page breaks in large align blocks

\usepackage[disable,textsize=tiny]{todonotes}
\icmltitlerunning{Fixing the Double Penalty in Data-Driven Weather Forecasting Through a Modified Spherical Harmonic Loss Function}

\begin{document}

\twocolumn[
\icmltitle{Fixing the Double Penalty in Data-Driven Weather Forecasting \\
           Through a Modified Spherical Harmonic Loss Function}




\begin{icmlauthorlist}
\icmlauthor{Christopher Subich}{mrd}
\icmlauthor{Syed Zahid Husain}{mrd}
\icmlauthor{Leo Separovic}{mrd}
\icmlauthor{Jing Yang}{mrd}
\end{icmlauthorlist}

\icmlaffiliation{mrd}{Meteorological Research Division, Environment and Climate Change Canada, Dorval, Quebec, Canada}

\icmlcorrespondingauthor{Christopher Subich}{christopher.subich@ec.gc.ca}

\icmlkeywords{Machine Learning, ICML, Numerical Weather Prediction, Spherical Harmonics}

\vskip 0.3in
]



\printAffiliationsAndNotice{}  % leave blank if no need to mention equal contribution

\begin{abstract}
    Recent advancements in data-driven weather forecasting models have delivered deterministic models that outperform the leading operational forecast systems based on traditional, physics-based models. However, these data-driven models are typically trained with a mean squared error loss function, which causes smoothing of fine scales through a ``double penalty'' effect.  We develop a simple, parameter-free modification to this loss function that avoids this problem by separating the loss attributable to decorrelation from the loss attributable to spectral amplitude errors.  Fine-tuning the GraphCast model with this new loss function results in sharp deterministic weather forecasts, an increase of the model's effective resolution from 1,250 km to 160 km, improvements to ensemble spread, and improvements to predictions of tropical cyclone strength and surface wind extremes.
\end{abstract}

\section{Introduction}\label{sec:intro}

% Rise of data-driven weather forecasting

The models developed in \citet{weyn_2020} and \citet{keisler_2022} suggested that deep neural networks might ``solve'' the problem of medium-range weather forecasting with data-driven machine learning models.  In 2023, the release of GraphCast \citep{graphcast}, FourCastNet \citep{fourcastnetv1}, and Pangu-Weather \citep{pangu} demonstrated forecast skill that met or surpassed that of the high-resolution forecast system (IFS) of the European Centre for Medium Range Weather Forecasts (ECMWF) at lead times (forecast lengths) up to 10 days, and some commenters \citep{future_ai} anticipated that data-driven forecasting would soon supplant traditional numerical weather prediction (NWP) in all operational contexts.  Since the publication of these models, the field has been joined by many others, including the Artificial Intelligence Forecasting System (AIFS) developed by ECMWF itself \citep{aifs}.

From the standpoint of machine learning, atmospheric forecasting is a large-scale generative problem comparable to predicting the next frame of a video.  As a typical example, the version of the GraphCast model deployed experimentally by the National Oceanic and Atmospheric Administration (NOAA) \citep{noaa_graphcast_tr,noaa_graphcast} predicts the 6-hour forecast for six atmospheric variables at each of 13 vertical levels plus five surface variables, on a ¼° latitude/longitude grid, for about 86 million output degrees of freedom in aggregate.  GraphCast takes two time-levels as input, so the input for this model has about 170 million degrees of freedom.

These first-generation data-driven weather models generally act as deterministic forecast systems, where each unique initial condition is mapped to a single forecast and verified against a ``ground truth'' from a data analysis system.  The ERA5 atmospheric reanalysis \citep{era5} of ECMWF is most often used as the source of initial and verifying data for these forecast systems owing to its high quality and consistent behaviour from 1979 to present.

\begin{figure}[ht]  
    \vskip -0.1in
    \begin{center}
    \centerline{\includegraphics[width=0.35\textwidth]{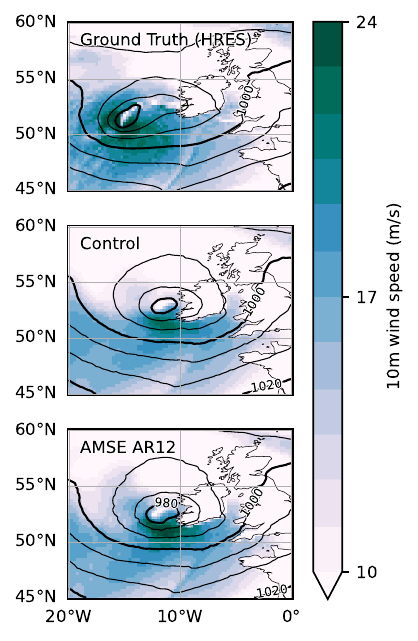}}
    \vskip -0.2in
    \caption{10 m wind speed and mean sea level pressure for winter storm Eunice, 18 Feb 2022 at 0 h UTC.  Top: HRES data at ¼° (ground truth), middle: 3.5d forecast produced by GraphCast, bottom: this work.  This work produces an overall sharper forecast, with a better prediction of the winter storm's strength.}
    \label{fig:smooth_forecast}
    \end{center}
    \vskip -0.4in
\end{figure}

\subsection{The Problem of Forecast Smoothing}

Despite their overall forecast skill, deterministic data-driven forecast systems are universally understood to produce overly-smooth forecasts.  A typical example of this behaviour is shown in figure \ref{fig:smooth_forecast} where a 3.5-day prediction of winter storm Eunice by the 13-level, ¼° GraphCast model is too weak and overly smooth.  This smoothing results in an under-prediction of localized extreme events, and it makes the model less suitable for downstream tasks such as spectral nudging \citep{nudging} and data assimilation \citep{noaa_enkf}.

This smoothing is most-discussed in relation to the prediction of gridded, global weather fields, but it is still present in models that have radically different architectures.  \citet{vaughan_2024} develops a model that operates directly in observation space without an underlying grid that still produces smooth forecasts of the future, and \citet{han_2024} shows diminished forecast activity (a bulk measure related to blurring) at longer lead times for a local-area model despite a nominal kilometer-scale resolution.

The conventional wisdom is that this smoothing is something that can be fixed in the context of an ensemble forecasting system, which produces realizations from the space of potential future forecasts.  GenCast \citep{gencast} and AIFS-CRPS \citep{aifs_crps} directly produce a stochastic ¼° forecast given initial values and a source of random noise.  SEEDS \citep{seeds} and ArchesWeatherGen \citep{archesweathergen} are examples of models that predict variations around an ensemble mean, using the generative step to ``fill in the blanks'' around a smooth baseline.  \citet{lippe_2023} approaches this problem from a more general partial differential equation framework, and it develops a diffusion method that iteratively refines finer scales.

Of these examples, all but AIFS-CRPS use a diffusion technique with mean squared error (MSE) used as the de-noising loss function, while AIFS-CRPS instead uses the continuous ranked probability score (CRPS, \citet{crps}) as its loss function to directly optimize the spread/error relationship of its produced ensemble.

However, we think that the problem of generating a good ensemble is distinct from the problem of forecast sharpness and effective resolution.  Traditional NWP systems try to directly model the physics of the atmosphere, such that the system's forecasts are always plausible atmospheric states without excessive smoothing.  Turning such a system into an ensemble prediction system involves supplying it with perturbed initial conditions and possibly stochastically perturbing the model's sub-grid parameterizations \citep{spp,spp2}.  

In the machine learning space, \citet{fcn_ensemble} develops a well-calibrated large ensemble using 29 independently-trained instantiations of the \citet{fourcastnetv2} architecture.  When combined with initial-condition perturbations, the result was a well-calibrated large ensemble, despite each individual ensemble member suffering from the smoothness typical of deterministic data-driven forecast systems.

\citet{lagerquist_can_2022} also develops a variety of loss functions based on the same spatial methods (such as filtering and max-pooling) to verify forecasts of convective events like thunderstorms in evaluation of high-resolution, limited-area models.

\subsubsection*{NeuralGCM}
    % NeuralGCM's zoo of loss functions, including sharpness
        % NeuralGCM as hybrid model has direct energy cascade

NeuralGCM \citep{neuralgcm} is one of the few global data-driven models that has addressed the problem of smoothing even in deterministic (non-ensemble) configurations.  However, this model is difficult to compare with its peers.  It has a hybrid architecture, combining a classical dynamical core with a learned network for sub-grid parameterizations that acts independently at each vertical column, and the classical dynamical core should cause fine-scale features to develop naturally.  In addition, the model was trained using a weighted sum of several loss functions, one of which uses MSE only on a coarsened (smoothed) version of the forecast and verifying analysis while another matches the spherical harmonic power spectrum (but not phase) only at high wavenumbers (short scales).  It is not clear which of these properties are necessary or sufficient to reduce the smoothing of deterministic NeuralGCM forecasts, and the use of several loss functions adds many degrees of freedom in their weighting and internal filtering.

\subsection{This Work}

The purpose of this work is to tackle the problem of smoothing in a purely deterministic, data-driven setting: can we produce a sharp forecast of the atmosphere without directly modelling ensemble uncertainty?  Our answer is ``yes.''  By modifying the MSE loss function to smoothly interpolate between amplitude-preservation and classical MSE, we can efficiently fine-tune a version of the GraphCast model to fix its smoothing problem and reproduce sharp forecasts.  This greatly increases the model's effective resolution, producing better predictions of tropical cyclone intensity and surface wind speed. 

Section \ref{sec:method} describes the modified loss function, its theory of operation, and the fine-tuning procedure used for this work.  Section \ref{sec:results} presents verification results of the fine-tuned GraphCast model, and section \ref{sec:discussion} concludes with discussion of the method's limitations and potential extensions.  Appendix \ref{app:mle} discusses the loss function in the context of maximum likelihood estimation, and appendix \ref{app:verif} presents more detailed verification statistics.

\section{Method}\label{sec:method}

\subsection{Smoothing Is Optimal Under Mean Squared Error} \label{subsec:optimal}
    % double-penalty
    % prediction of the ensemble mean
    % Smoothing corrupts headline statistics: comparison of results for DM¼°?
    % Smoothing over training for 1° Graphcast

In the NWP community, model evaluation using the mean squared error is widely understood to suffer from a so-called ``double penalty'' \citep{double_penalty1,double_penalty2}.  Under MSE, a good forecast that correctly predicts a feature such as a storm but misses its location is penalized twice compared to a perfect forecast, once for missing the storm at its correct location and again for predicting a storm at an incorrect location.  In traditional NWP, this double penalty makes model verification more difficult, particularly when studying the impact of improvements to forecast resolution that create more opportunities for misplaced predictions.  

When MSE is used as the loss function to train a data-driven model, the double penalty problem is more than annoyance: it encourages the model to generate unrealistically smooth predictions by reducing the amplitude of unpredictable scales.  To show this quantitatively, consider the case of predicting a single variable.  Let $Y = \normal(0,1)$ be the target, and let $X$ be the imperfect prediction of that target, modelled as a normal random variable with a standard deviation of $\sigma_X=\sqrt{\expect(X^2)}$ and correlation coefficient of $\rho=\expect(XY)/\sigma_X$, where $\expect(\cdot)$ is the expectation operator.  Writing $X$ in terms of a correlated and an uncorrelated component gives:
\begin{equation} \label{eqn:corr1}
    X = \sigma_X ( \rho Y + \sqrt{1-\rho^2} \normal(0,1)),
\end{equation}
and the corresponding expected MSE is:
\begin{align} \label{eqn:mse1}
    \expect{(\mse(X,Y))} &= \expect((X-Y)^2) \nonumber \\
                        &= \expect(X^2) + \expect(Y^2) - 2\expect(XY) \nonumber \\
                        &= \sigma_X^2 + 1 - 2\sigma_X \rho.
\end{align}
For fixed $Y$, this MSE is optimized with a perfect prediction, when $\sigma_X = 1$ and $\rho=1$.  However, if $0 < \rho < 1$ because the process is only partially predictable, the MSE is optimized with respect to $\sigma_X$ when $\sigma_X=\rho < 1$, leading to an underprediction of the process's natural variability.

\subsection{Spectral Separation of the Mean Squared Error}

Predictions of global weather are high-dimensional, but equations \eqref{eqn:corr1} and \eqref{eqn:mse1} can be extended to any decomposition (partition of unity) of the prediction and target fields that obeys Parseval's theorem.  Taking this decomposition point-by-point, extending the analysis to include a nonzero mean, and taking the expectation over an ensemble of predictions gives rise to skill/spread evaluations.  However, this decomposition is not possible at training time for a deterministic data-driven weather forecast, and instead we turn to a spherical harmonic decomposition.

Let $\harm_k^l(i,j)$ be the complex-valued spherical harmonic mode with total wavenumber $k$ and zonal wavenumber $l$ at the $(i,j)$ grid point on a latitude/longitude grid, normalized such that $\int \harm_k^l (\harm_m^n)^* = \delta_{km} \delta_{ln}$, where $(\cdot)^*$ is the complex conjugate\footnote{In practice, this work takes advantage of the property that $\harm_k^{-l} = (\harm_k^l)^*$ to work with only non-negative wavenumbers.}.  A scalar field $x(i,j)$ defined on the latitude/longitude grid can be written in terms of spherical harmonics as:
\begin{equation*}
    x(i,j) = \sum_k \sum_{l=-k}^{k} \alpha_x(k,l) \harm_k^l(i,j),
\end{equation*}
with $\alpha_x(k,l)$ the corresponding spectral coefficient.  For two fields $x$ and $y$ the latitude-weighted MSE is:
\begin{align} \label{eqn:smse_prelim}
    \mse(x,y) &= \sum_i \sum_j \mathrm{dA}(i,j) (x(i,j) - y(i,j))^2 \nonumber \\
             &= \sum_k \sum_{l=-k}^{k} \lvert \alpha_x(k,l) - \alpha_y(k,l) \rvert^2,
\end{align}
where the $\mathrm{dA}$ term is incorporated into the normalization of $\harm_k^l$.  Importantly, $\alpha_x$ and $\alpha_y$ are independent with respect to zonal and total wavenumber, but the double summation here now allows us to group these terms in a physically meaningful way.  Grouping terms in the inner (zonal) sum together gives rise to the power spectral density $\psd_k(x) = \sum_l |\alpha_x(k,l)|^2$ and coherence $\coh_k(x,y) = \sum_l \real{(\alpha_x(k,l) \alpha_y^*(k,l))} / \sqrt{\psd_k(x) \psd_k(y)}$ (where $\real{(\cdot)}$ takes the real part) as scale-dependent analogs to variance and correlation respectively.  Performing the appropriate substitutions:
\begin{align}\label{eqn:smse}
    \mse(x,y) &= \sum_k PSD_k(x) + PSD_k(y) - \nonumber \\
             & 2 \sqrt{PSD_k(x) PSD_k(y)} \coh_k(x,y).
\end{align}
If $x$ is taken to be a forecast field and $y$ is the ground-truth analysis, as in \eqref{eqn:mse1} this is minimized when $\sqrt{PSD_k(x) PSD_k(y)^{-1}} = \coh_k(x,y)$

\begin{figure}[tb]
    \vskip -0.1in
    \begin{center}
    \centerline{\includegraphics[width=0.4\textwidth]{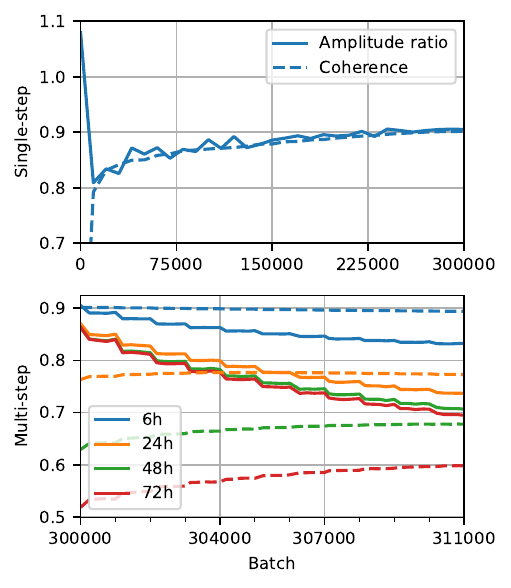}}
    \vskip -0.2in
    \caption{Amplitude ratio (solid) and coherence (dashed) for the spherical harmonic mode with total wavenumber 100 for temperature at 850hPa during the training of a 1° version of the GraphCast model with an MSE loss function.  At top, values for 6h lead time during the single-step pre-training phase and at bottom, values for 6h--72h during the forecast rollout (batches 300,000--311,000, incrementing one step every 1,000 batches).}
    \label{fig:1deg_smoothing}
    \end{center}
    \vskip -0.4in
\end{figure}

This optimum leads to the observed smoothing in data-driven models through two factors:
\begin{itemize}
    \item Fine scales (large $k$, short wavelengths) are generally less predictable than coarse scales (small $k$, large wavelengths), particularly at longer lead times, and
    \item Data-driven models with conventional architectures learn to smooth fine scales (reducing the power spectral density) more quickly than they learn to predict them (increasing coherence).
\end{itemize}
This is illustrated in figure \ref{fig:1deg_smoothing}, which shows the amplitude ratio (square root of power spectral density ratio) and coherence for total wavenumber 100 (wavelength about 400 km) between predictions of the temperature field at the 850hPa level and the ground truth, for a 1° version of GraphCast during training with the curriculum of \citet{graphcast}.  After a rapid adjustment from initially random outputs, the amplitude ratio and coherence closely track each other, with an initial smoothing followed by a gradual but partial sharpening as the model learns to predict this scale (with increasing coherence).  When training is extended autoregressively to 12 steps (72h forecasts), smoothing increases at longer lead times as the forecast length increases.

\subsection{Spectrally Adjusted Mean Squared Error}

This smoothing is undesirable.  It makes the produced forecasts less realistic, and it complicates model comparisons.  \citet{graphcast} performs extensive verification under an ``optimal blurring'' model to show that the purported forecast power of GraphCast is not just an artifact of its smoothing, and more straightforward verification methodologies such as that of \citet{weatherbench2} may conflate the effects of more-optimal smoothing with forecast skill even when evaluating at reduced resolution.  It would instead be far more desirable if the loss function reflected our true goal, encouraging forecasts to correlate well to the ground-truth and retain realistic variation at finer scales.

Fortunately, beginning with MSE written in terms of its spectral decomposition, this is a simple modification.  First, we write \eqref{eqn:smse} in terms of a perfectly-correlated loss (with $\coh_k(x,y)=1$) and a residual:
\begin{align}
    \mse(x,y) &= \sum_k (\sqrt{\psd_k(x)} - \sqrt{\psd_k(y)})^2 + \nonumber \\
              2 &\sqrt{\psd_k(x)\psd_k(y)} (1-\coh_k(x,y)). \label{eqn:smse_decomp}
\end{align}
Then, we seek to break the interaction between the spectral amplitudes and coherence contained in the second term of \eqref{eqn:smse_decomp}.  One option would be to fix the role of $x$ as a trial prediction and $y$ as the verifying analysis and replace $\sqrt{\psd_k(x) \psd_k(y)}$ by $\psd_k(y)$, but the symmetry of the loss function can be retained by writing:
\begin{align}
    \amse(x,y) &= \sum_k (\sqrt{\psd_k(x)} - \sqrt{\psd_k(y)})^2 + \nonumber \\
                2 \max(&\psd_k(x),\psd_k(y)) (1-\coh_k(x,y)). \label{eqn:amse}
\end{align}
AMSE is now an adjusted mean squared error, which can act as a drop-in replacement during model training.  Like its unmodified counterpart, AMSE is zero if and only if $x=y$, and it has the same Taylor expansion (in $x$) about $x=y$.  The gradients of $-\amse(x,y)$ with respect to $x$ (that is, minimizing AMSE) will always point in the direction of increased coherence ($\coh_k(x,y) \to 1$) and a correct spectral magnitude ($\psd_k(x) \to \psd_k(y)$), even if physical limits to predictability impose a practical limit to coherence.  AMSE retains the units of MSE and has a similar magnitude, but it is no longer a proper metric because it does not satisfy the triangle inequality. %\footnote{As an example, $\amse(x,-x) > \amse(x,\epsilon x) + \amse(\epsilon x, -\epsilon x) + \amse(-\epsilon x, -x)$ for small $\epsilon$.}

Unlike the mix of filtered and spectral loss functions used by NeuralGCM, \eqref{eqn:amse} is parameter-free, requiring no selection of cutoff scales or scaled addition of qualitatively different terms.  A parameter could be added to \eqref{eqn:amse} to change the relative weights of its two terms, but that was not necessary in this work.  Appendix \ref{app:mle} contemplates extending this framework to maximum likelihood estimation.

Equation \ref{eqn:amse} is defined for a single two-dimensional variable, but GraphCast produces several outputs per gridpoint.  In the ¼°, 13-level version of the model considered here, there are six variables (geopotential, temperature, specific humidity, two components of horizontal wind, and vertical wind) produced at each of 13 atmospheric levels plus five variables (2-meter temperature, two components of 10-m horizontal wind, mean sea level pressure, and 6h-accumulated precipitation) at the surface.  This work follows equation (A.19) of \citet{graphcast} by aggregating each variable's error (MSE there, AMSE here) with a per-variable weight, level weighting proportional to the pressure level, and normalization of the disparate units by a per-variable, per-level standard deviation.

\begin{table}[tb]
    \caption{Fine-tuning curriculum for the ¼°/13-level version of GraphCast trained for this study, including the peak and terminal learning rates (LR) of the cosine annealing schedule used at each stage.  The batch size was 8 throughout, and each stage had a warm-up period of 64 batches.} \label{tab:finetune}
    \vskip 0.10in
    \begin{center}
    \begin{small}
    \begin{tabular}{cccc}
        \toprule
        Length & Batches & Peak/End LR & GPU Time \\ 
        \midrule 
        1 step (6h)    & 25,000 & $2.5 \cdot 10^{-5}$/$1.25 \cdot 10^{-7}$ & 7.7d \\
        2 steps (12h)  & 2,500  & $2.5 \cdot 10^{-6}$/$7.5 \cdot 10^{-8}$  & 2.2d \\
        4 steps (24h)  & 2,500  & $2.5 \cdot 10^{-6}$/$7.5 \cdot 10^{-8}$  & 4.3d \\
        8 steps (48h)  & 1,250  & $2.5 \cdot 10^{-6}$/$7.5 \cdot 10^{-8}$  & 4.6d \\
        12 steps (72h) & 1,250  & $2.5 \cdot 10^{-6}$/$7.5 \cdot 10^{-8}$  & 7.4d \\
        \bottomrule
    \end{tabular}
    \end{small}
    \end{center}
    \vskip -0.15in
\end{table}

\subsection{Fine-Tuning Methodology}
% Fine-tuning methodology
    % HRES dataset, Subich preprint
    % Separate evaluation for AR1 and AR12

We demonstrate the efficacy of this loss function using a ¼°, 13-level version of GraphCast.  Based on the observation above that the model tends to rapidly adjust its per-scale smoothing to match its coherence, we treat this as a fine-tuning process and begin with the ``operational'' checkpoint provided by \citet{graphcast_github}, which is publicly available under a Creative Commons license.

Our fine-tuning methodology is summarized in table \ref{tab:finetune}, and the overall approach is inspired by \citet{subich_gc_gdps}.  While the baseline model checkpoint was trained over 72h (12 autoregressive steps of 6h each), in earlier testing at 1° we found it better to begin the fine-tuning with single-step forecasts and increase the forecast length in stages.  Training over single steps is both faster per step and supports higher learning rates.

The other training hyperparameters, including AdamW \citep{adamw} hyperparameter settings and per-variable, per-level loss weightings were identical to those described by \citet{graphcast}.%  Replacing MSE with AMSE in the training process was a simple matter of replacing the ``spatial location'' summation and grid area-weight in its equation (19) with our \eqref{eqn:amse}.

The 13-level GraphCast checkpoint that forms the base of our fine-tuned model was originally trained on the ERA5 reanalysis from 1979--2017, then itself fine-tuned on the initial conditions used for the contemporaneous HRES (IFS) model from ECMWF over 2016--2021.  We used this latter dataset and training period in our work, and it is available from \citet{weatherbench2} as the ``HRES-fc0'' dataset.  As described in \citet{graphcast}, we supplemented the HRES data with the accumulated precipitation field from the ERA5 reanalysis over the training period, since an initial conditions dataset has no accumulated precipitation by definition.  The equivalent data for calendar year 2022 is also available, and we used this period for model evaluation.

We fine-tuned our model on 1-2 nodes of a cluster containing 4 NVidia A100 40GiB GPUs per node, using data parallelism with the batch split across GPUs and the gradients accumulated via MPI.  Overall, the fine-tuning process took about 26.2 GPU-days.  \citet{graphcast} does not disclose the total training time required to produce the model checkpoint from scratch, but other models of similar size \citep{fourcastnetv2,aifs,pangu} report training times of about 1 GPU-year using similar hardware.

\begin{figure}[t!]
    \vskip -0.1in
    \begin{center}
    \centerline{\includegraphics[width=0.375\textwidth]{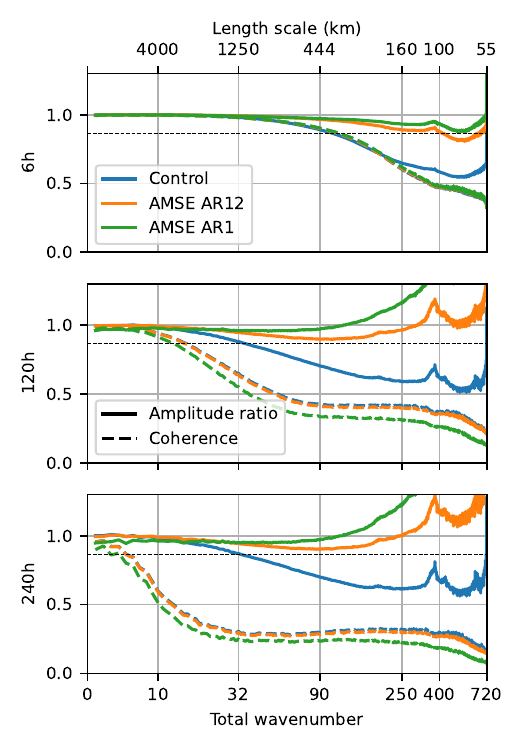}}
    \vskip -0.2in
    \caption{Amplitude ratio (solid) and coherence (dashed) for all output variables and levels, weighted using the variable/level weights in the loss function, for the control model and this work after the 1-step training and after complete fine-tuning.  Top: 6h lead time, middle: 120h (5d) lead time, bottom: 240h (10d) lead time.  The dashed line is placed where a model would underrepresent the power spectral density by 25\%. %The control model smooths out fine scales, particularly at long lead times, while after fine-tuning this problem is largely eliminated.
    }
    \label{fig:resolution}
    \end{center}
    \vskip -0.4in
\end{figure}

\section{Results}\label{sec:results}

The fine-tuned model is evaluated against the control (unmodified) model over calendar year 2022 using the HRES dataset for initialization and as ground truth unless otherwise specified.  As reported in \citet{graphcast} and is typical in other deterministic data-driven models, forecast performance at longer lead times improves when the model is autoregressively trained over multiple steps, and the fully-tuned model (trained over 12 forecast steps and labelled ``AMSE AR12'' in the figures and discussion below) is considered the primary model for evaluation.

Since multi-step training also tends to cause both fine-scale smoothing and a loss of variability in ensemble settings, these respective evaluations (sections \ref{sec:resolution} and \ref{sec:ensemble}) will also include the model checkpoint created after just single-step fine-tuning, denoted ``AMSE AR1.''

\begin{figure*}[t]
    \vskip 0.1in
    \begin{center}
    \centerline{\includegraphics[width=0.8\textwidth]{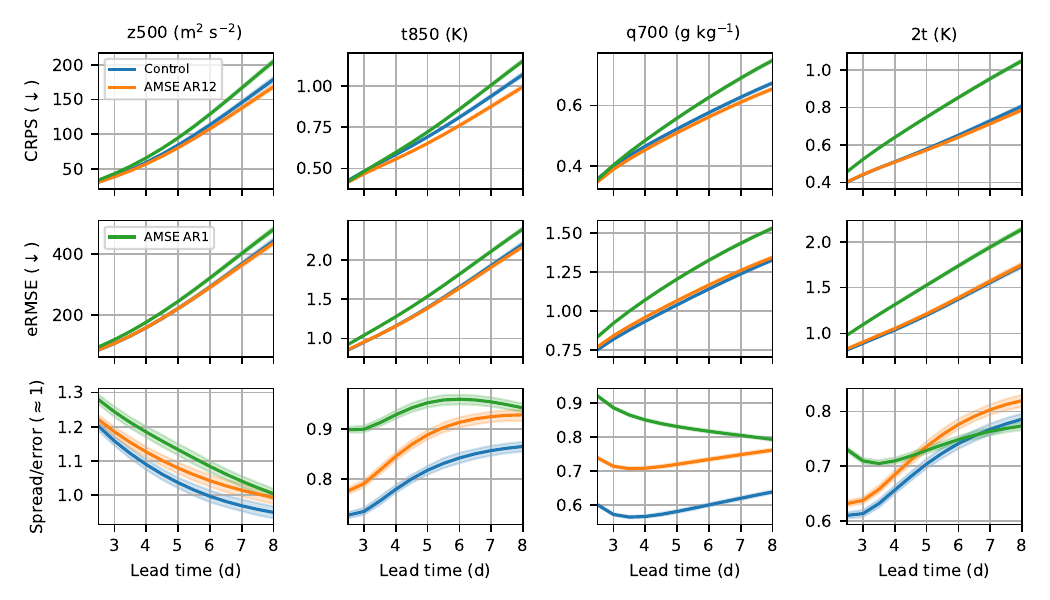}}
    \vskip -0.2in
    \caption{Lagged ensemble statistics for geopotential (z) at 500hPa, temperature (t) at 850hPa, specific humidity (q) at 700hPa, and 2-meter temperature (2t) from left to right.  The statistics are the CRPS, root mean squared error of the ensemble mean, and spread-error ratio, from top to bottom.}
    \label{fig:ens_stats}
    \end{center}
    \vskip -0.4in
\end{figure*}

\subsection{Effective Resolution}\label{sec:resolution}

Conventional, physics-based NWP models are widely understood to have an effective resolution that is coarser than the model's native grid resolution.  Limits to effective resolution come from the limited fidelity of spatial or temporal discretization, from artificial diffusion or damping used to stabilize a model, and from sub-grid processes (such as turbulence) that must be imperfectly estimated rather than directly modelled.  A model behaves unrealistically at scales finer than its effective resolution, typically providing insufficient variability and too-smooth solutions.

Deterministic data-driven NWP models do not have the same underlying numerical issues that result in reductions to effective resolution, but the smoothing produced by training with an MSE-based loss function acts in a very similar way.  Figure \ref{fig:resolution} shows the amplitude ratio ($\sqrt{\psd_k(x) \psd_k^{-1}(y)}$) and coherence ($\coh_k(x,y)$) between each of the GraphCast models and the verifying analysis over calendar year 2022.  To compute a combined curve despite the many per-gridpoint values predicted by the model, the statistics for each separate variable are combined using the same variable and level weighting used in the model's loss function\footnote{Normalization of the disparate variables by standard deviation was not required here, since the amplitude ratio and coherence are already dimensionless.}.

The control model significantly smooths fine scales even after a single 6-hour forecast step, and that smoothing increases with the forecast lead time.  If we somewhat arbitrarily draw the line of effective resolution at the point where the model has lost 25\% of the per-wavenumber energy (corresponding to a ratio of power spectral densities of $0.75$ or an amplitude ratio of $\sqrt{0.75}$), the 5-day predictions of the control model reach that cutoff at wavenumber 32, corresponding to oscillations with a wavelength of about 1250 km.  Small changes in the target amplitude ratio will result in small changes to the derived effective resolution.

The models fine-tuned in this work do not show this type of fine-scale dissipation.  The AMSE AR12 model has a small amount of smoothing at moderate scales, but the variability recovers again at finer scales, and a dissipation-based definition of effective resolution would be extremely sensitive to the cutoff value.  Instead, we observe that for longer forecasts the model has more energy at small scales than in the ground-truth dataset, suggesting a ``noise-based'' definition of effective resolution.  For long forecasts, the amplitude ratio rises above 1 around wavenumber 250, giving an effective resolution of about 160 km.  

The AMSE AR1 model shows the same qualitative behaviour but generates this ``noise'' more strongly, leading to a reduced effective resolution of about 450 km (wavenumber 90).  The forecasts produced by this version of the model are less coherent with the analysis, showing a reduced forecast skill at all scales for longer forecasts.

For illustration, appendix \ref{app:spec} shows amplitude spectra for select variables at various lead times, without normalizing by the spectral magnitude of the ground truth.  Appendix \ref{app:ablation} discusses the effective resolution of the model when trained with either mean squared error or mean absolute (L1) error.

\subsection{Lagged Ensemble Verification}\label{sec:ensemble}

The observation that AMSE-based fine-tuning provides sharp forecasts is encouraging, but that alone is not enough to demonstrate utility.  The model might have learned to match its expected variance by generating quasi-static noise that does not sufficiently depend on the surrounding flow, for example.  The ideal way to measure this sort of forecast skill is in an ensemble setting, where the chaotic nature of the atmosphere is accounted for by evaluating the full distribution of plausible outputs given an initial condition.

Development of a full ensemble system is well beyond the scope of this work, but \citet{lagged_ensemble} provides a procedure to evaluate a deterministic model using an ensemble generated from time-separated initial conditions.  The central idea of this method is that predictions initialized at different times should diverge, so several consecutively-initialized forecasts that are all valid at a shared time form an ad-hoc ensemble, without the need for an auxiliary method of defining an ensemble of initial conditions.

This approach is implemented here, using forecasts initialized at 12-hourly intervals in 2022 and evaluated from 10 January 2022 0:00 UTC to 31 December 2022 12:00 UTC. Each set of nine consecutively initialized forecasts (spanning four days from beginning to end) forms an ensemble, and the ensemble's notional lead time is that of its central member.  

The primary evaluation metrics are the CRPS, ensemble root mean squared error (eRMSE), and spread/error ratio, with definitions given in appendix \ref{app:ensemble}.  For an operational ensemble, a spread/error ratio close to 1 is considered ideal, but that is confounded here because the members of a lagged ensemble are not statistically interchangeable.  Since deterministic data-driven NWP models are underdispersive, however, a larger spread/error ratio is generally better.

\begin{figure}[tb]  
    \vskip 0.1in
    \begin{center}
    \centerline{\includegraphics[width=0.35\textwidth]{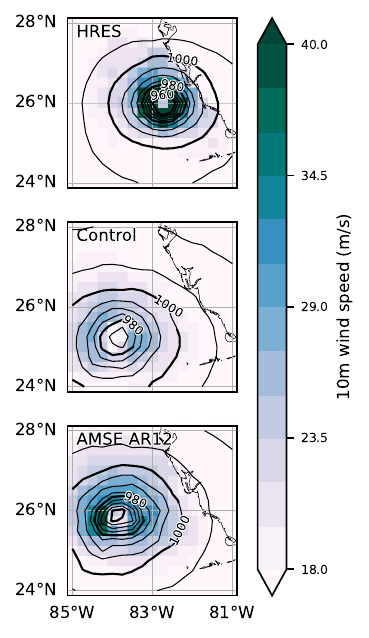}}
    \vskip -0.2in
    \caption{10 m wind speed and mean sea level pressure for Hurricane Ian, 28 Sept 2022 at 12h UTC.  Top: HRES data at ¼°, middle: 5d forecast produced by the control GraphCast model, bottom: the model after 12-step fine-tuning with AMSE.}
    \label{fig:hurricane_ian}
    \end{center}
    \vskip -0.4in
\end{figure}

Figure \ref{fig:ens_stats} shows the evolution of these statistics versus lead time for a selection of variables and levels, and more detailed evaluation of CRPS and eRMSE are shown in figures \ref{fig:crps_scorecard} and \ref{fig:ermse_scorecard}.  The AMSE AR12 model shows consistent improvements to the CRPS while the eRMSE sees little change, indicating that the fine tuning process produces a better-calibrated (more dispersive) ensemble without degrading overall predictive performance.  

While the AMSE AR1 model shows greater ensemble spread, the less skillful forecast results in a significantly reduced CRPS.  However, unlike the results of \citet{lagged_ensemble}, the spread/error ratio of the AR1 and AR12 models converge for most variables at longer lead times, suggesting that multi-step training in this framework does not cause a collapse of variability in an ensemble setting.

\begin{figure}[tb]
    \vskip 0.1in
    \begin{center}
    \centerline{\includegraphics[width=0.45\textwidth]{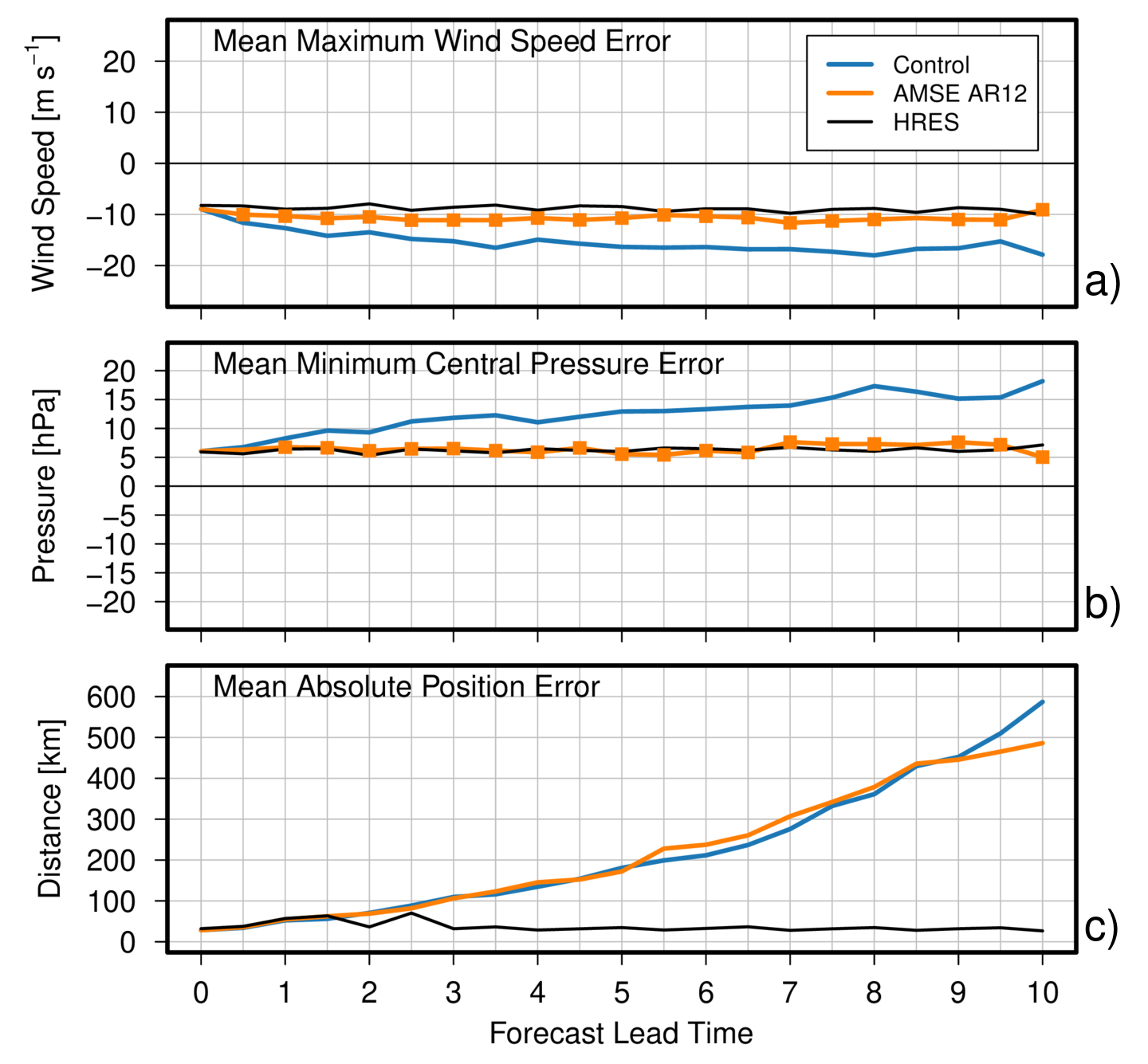}}
    \vskip -0.2in
    \caption{Predictions of tropical cyclone intensity ((a), mean maximum surface wind speed; (b) mean minimum central pressure) and mean absolute position error (c) for forecasts initialized 20 June--19 September 2022.  Orange squares show statistically significant differences between the AMSE AR12 and control predictions.}
    \label{fig:tc_stats}
    \end{center}
    \vskip -0.4in
\end{figure}

\subsection{Hurricane Prediction and Extreme Weather}
% Hurricane test case: Ian

The effect of improved effective resolution is most strongly apparent in the prediction of local extremes, and few weather events are more extreme than tropical cyclones.

Data-driven NWP models like GraphCast improve predictions of hurricane tracks relative to conventional NWP models (see for example figure 3A of \citet{graphcast}).  Since storms are guided by large-scale ``steering flows'' that have natural scales of thousands of kilometers, these predictions of storm position are relatively unaffected by the models' limited effective resolutions but benefit from improvements in large-scale forecast skill.  However, cyclones themselves are comparatively small, and predictions of the storm intensity are significantly affected by MSE-induced smoothing.

Figure \ref{fig:hurricane_ian} depicts this situation for Hurricane Ian, the most intense Atlantic tropical cyclone of the 2022 season.  Both the control version of GraphCast and the AMSE AR12 version produce a reasonable 5-day prediction of the storm's location (within about 125 km), but the control version of GraphCast predicts an unrealistically weak storm.

% Hurricane bulk validation

More quantitatively, figure \ref{fig:tc_stats} shows the mean intensity and mean absolute position errors for tropical cyclones over 20 June--19 September 2022 initializations, using the algorithm of \citet{zadra_evaluation_2014} to compare against the International Best Track Archive for Climate Stewardship database \citep{ibtracs}.  Compared to these observations, even the HRES data is imperfect and shows a weak-intensity bias.  The control model has a larger weak-intensity bias that increases with lead time, but the AMSE AR12 model retains the quality of the HRES dataset.  The storm location predictions between the control and AMSE AR12 models are equivalent.

%Compared to these observations even the HRES data is imperfect, but the control model still shows a further weak bias that increases with lead time.  In contrast, the AMSE AR12 model has essentially no intensity bias compared to the HRES dataset, and its storm location predictions are equivalent to those of the control model.

% QQ plots for surface station temperature, wind speed

Extreme weather includes more than tropical cyclones, and appendix \ref{app:qq} discusses quantile-quantile predictions of surface wind speed and temperature, validated against station observations.  Both the control model and AMSE AR12 produce realistic temperature extremes, but the AMSE AR12 model provides more realistic predictions of wind-speed extremes. 

\section{Discussion \& Limitations}\label{sec:discussion}

Using the mean squared error as a model loss function asks the model to average away unpredictable scales.  In weather forecasting, the unpredictable scales are generally the smaller scales that carry information about local variance, and this averaging process leads to data-driven weather forecasts that are far smoother than the grid resolution would suggest.  

This is not a property inherent to data-driven NWP.  The alternate loss function based on \eqref{eqn:amse} uses a spectral transform to separate the loss attributable to amplitude error from that attributable to decorrelation, encouraging the model to reproduce a realistic spectrum even if it can't make an accurate prediction.  When applied to the ¼°, 13-level version of GraphCast with an abbreviated fine-tuning process, we recover a model that has a much finer effective resolution, has improved CRPS-based verification in a lagged ensemble setting, and fixes the weak intensity bias in the prediction of tropical cyclones.

When fine-tuned autoregressively over multiple forecast steps, the model suffers from a small amount of smoothing at mesoscales (intermediate scales).  We speculate that this is because such autoregressive training has two objectives: forecasts are asked both to be accurate (and thus sharp, per \eqref{eqn:amse}) and to be good initial conditions for the next forecast step.  This latter goal is implicit, and it is not directly affected by the loss function used in training.  Future work will consider the use of a replay buffer in training (like that of \citet{fengwu}) to see if long-range forecast skill might be retained with even better prediction of amplitudes.

Since the AMSE loss function \eqref{eqn:amse} is zero if and only if the predicted field matches the ground truth, it may be useful throughout model training rather than just during a fine-tuning pass.  However, a thorough  test of this proposition would require a considerable computational budget, so it is left for future work.  Use of the AMSE loss function throughout the training process might improve the coherence of fine-scale prediction by allowing the model to spend more of its training time ``seeing'' these modes, but on the other hand the coherence-dependent smoothing encouraged by the MSE loss function (figure \ref{fig:1deg_smoothing}) might act as an implicit regularization that smooths the model's gradients and speeds up training overall.

\subsection{Effective Resolution}
The ultimate conclusion of this work is that the AMSE-based error measure improves the effective resolution of NWP weather models, but the phrase ``effective resolution'' must always be accompanied by the question, ``effective at what?''

We chose to define an effective resolution based on smoothing of fine scales, since a model that simply doesn't represent a scale cannot effectively model it.  However, other definitions exist in the literature, and users of these models should keep their ultimate goals in mind.  For example, \citet{kent_2014} studies various discretization schemes for numerical partial differential equations under both diffusion (smoothing) and dispersion (wave propagation) definitions of effective resolution.

\subsection{Alternative Grids}

Passing from equation \eqref{eqn:mse1} to \eqref{eqn:smse} makes use of Parseval's theorem to give an exact relationship between the spatially-defined mean squared error and the equivalent in the spectral representation.  Implementing this in a training cycle requires fast computation of spherical harmonic transforms.  This is simple enough for global latitude/longitude grids, but it might be difficult for local-area models without a regular global grid structure. 

In these cases, we think that the basic intuition behind \eqref{eqn:amse} might still apply through other multiscale decompositions such as wavelet lifting \citep{sweldens}, provided suitable equivalents to scale-dependent variance and correlation could be found.  The multiscale decomposition is critical in some form, however, since the method takes advantage of the approximate independence of scale-separated modes.  Without such a decomposition (e.g. applying the adjustment of \eqref{eqn:amse} globally, without the harmonic transform), the model might be able to ``cover up'' a lack of fine-scale variability by over-emphasizing coarser scales.

\subsection{Applications to Other Domains}

AMSE is a natural error function for weather prediction because the spectral decomposition is physically meaningful and relatively stable over time.  A partially incorrect but realistic prediction of weather at 2000 km scales would not significantly change the amount of energy present at 100 km scales, just its relative location.  The goal of a deterministic forecast is to be physically plausible, and a correct prediction of spectral amplitudes is a necessary condition for physical plausibility.

The method can be mechanically applied whenever a spectral decomposition is possible, but additional value is only likely when a sub-aggregation of that spectrum is meaningful.  This is most obviously possible in other areas of fluid dynamics, particularly the modelling of turbulent flows.  In that domain, \citet{chakraborty_2025} developed a binned spectral loss function (on a planar domain) that is reminiscent of the amplitude-only component of \eqref{eqn:amse}, but it discards the phase information.  We are optimistic that integrating the spectral correlation along the lines of AMSE will make such models more robust.

\subsection{Applications to Ensemble Modelling}

We are particularly encouraged by the beneficial impact that AMSE-based training has on the spread of forecasts in an ensemble setting.  Without any dedicated ensemble-based training we end up with a model that nonetheless produces a more realistic spread of forecasts.  In future work, we hope to use this loss function as a basis for an ensemble forecast where each individual ensemble member produces a realistic trajectory, in addition to the whole-ensemble optimization encouraged by CRPS-like ensemble training.

\section*{Code and Data Availability}
An implementation of the AMSE error function and the code used to train GraphCast for this work are available at \url{https://github.com/csubich/graphcast/tree/amse} under the Apache 2.0 license.  The fine-tuned checkpoints produced for this study are available at \url{https://huggingface.co/csubich/graphcast_amse} under the CC-BY-ND-SA 4.0 license, as derivative works of the DeepMind ``graphcast-operational'' checkpoint.

\section*{Acknowledgements}

The authors would like to thank Charlie Hébert-Pinard, Vikram Khade, and Hugo Vandenbroucke-Menu of the Canadian Centre for Meteorological and Environmental Prediction for access to the 1\textdegree-trained version of GraphCast used to produce the results of figure~\ref{fig:1deg_smoothing}.

% \textbf{Do not} include acknowledgements in the initial version of
% the paper submitted for blind review.

% If a paper is accepted, the final camera-ready version can (and
% usually should) include acknowledgements.  Such acknowledgements
% should be placed at the end of the section, in an unnumbered section
% that does not count towards the paper page limit. Typically, this will 
% include thanks to reviewers who gave useful comments, to colleagues 
% who contributed to the ideas, and to funding agencies and corporate 
% sponsors that provided financial support.

\section*{Impact Statement}

Accurate weather forecasts are a vital public service, and its benefits are disproportionately concentrated in the extremes.  Accurate forecasts of extreme weather such as tropical cyclones save lives.  On one hand, this means that we should be eager to develop improvements to weather forecasting systems, but on the other hand it means that we should be very careful not to just ``chase scores,'' confusing what's easy to calculate with what's truly important.

This work contributes to this field by introducing a way to make data-driven weather forecasting more realistic, with variability at moderate and fine scales that is much closer to reality.  This improves various probabilistic scores and predictions of tropical cyclone intensity, but this is not a guarantee of complete physical plausibility.  In particular, we have not yet shown that these forecasts are better-behaved ``out of distribution,'' such as when simulating possible future climate paths.

Operational weather centres are very diligent about performing rigorous evaluation of models before making them operational, and we hope that this work can help ease the path towards the adoption of better-performing, data-driven forecasting systems in the near future.


\bibliography{paper_ml}
\bibliographystyle{icml2025}

%%%%%%%%%%%%%%%%%%%%%%%%%%%%%%%%%%%%%%%%%%%%%%%%%%%%%%%%%%%%%%%%%%%%%%%%%%%%%%%
%%%%%%%%%%%%%%%%%%%%%%%%%%%%%%%%%%%%%%%%%%%%%%%%%%%%%%%%%%%%%%%%%%%%%%%%%%%%%%%
% APPENDIX
%%%%%%%%%%%%%%%%%%%%%%%%%%%%%%%%%%%%%%%%%%%%%%%%%%%%%%%%%%%%%%%%%%%%%%%%%%%%%%%
%%%%%%%%%%%%%%%%%%%%%%%%%%%%%%%%%%%%%%%%%%%%%%%%%%%%%%%%%%%%%%%%%%%%%%%%%%%%%%%
\newpage
\appendix

\onecolumn

\section{Relationship to Maximum Likelihood Estimation}\label{app:mle}

In developing the AMSE loss function, the transformation from ordinary, gridpoint-based MSE \eqref{eqn:mse1} to its spectral definition with power spectral densities and coherence \eqref{eqn:smse_decomp} is algebraic in nature.  The beneficial effect of the AMSE loss function's separation of spectral-ampltiude and decoherence terms arises because the underlying spectral decomposition is physically meaningful.  At fine enough scales, atmospheric dynamics are increasingly rotationally symmetric and position-invariant, with individual spectral amplitudes that look like draws from a Gaussian distribution.

If we elevate this property from a fortunate coincidence to a simplifying assumption, we can treat the set of modes corresponding to a particular total wavenumber as random variables and apply the machinery of ensemble verification to individual, deterministic forecasts.  The goal of producing realistic forecasts despite limited predictability is conceptually similar to the goal of maximum-likelihood estimation, so we consider here the effect of Kullback-Leibler (KL) divergence minimization.  In the meteorology literature, the KL divergence is named the continuous ignorance score \citep{cis}, and it is sometimes used for ensemble verification.

Treat the modes corresponding to a single total wavenumber $k$ as a draw from a $2k-1$-dimensional normal random variable\footnote{That is, $k$ independent complex-valued modes from $1 
\ldots k$ with independent real and imaginary parts and a single, real zero-wavenumber mode.} with mean zero and some finite standard deviation.  In this interpretation, the ground-truth analysis is:
\begin{equation}
    Y_k = \sigma_Y \normal^{2k-1}(0,1).
\end{equation}
The forecast is itself taken to be a normal random variable, but following the pattern of \eqref{eqn:smse_decomp} it is partially correlated to $Y$ and has its own standard deviation. Take the correlation to be $\rho$ and the forecast standard deviation to be $\sigma_X$, and:
\begin{equation} \label{eqn:prob_x}
    X_k = \sigma_X \left( \frac{\rho}{\sigma_Y} Y + \sqrt{1-\rho^2} \normal^{2k-1}(0,1) \right),
\end{equation}
noting for emphasis that this definition of $X$ depends upon $Y$.  With the assumption that each of the per-wavenumber modes are independently drawn from this distribution, we can also treat $X$ and $Y$ as a product of $2k-1$ independent, scalar random variables, which will simplify the following algebra.

The KL divergence of the data given the forecast is then given by:
\begin{equation} \label{eqn:dkl1}
    \dkl(Y \Vert X) = \int P_Y(y') \log \left( \frac{P_Y(y')}{P_X(y')} \right) \mathrm{d}y',
\end{equation}
for the respective probability density functions (PDFs) $P_Y$ and $P_X$ and integrating over the space of possible observations parameterized by $y'$.  With these formulations, the PDF of $Y$ is simple:
\begin{equation} \label{eqn:py}
    P_Y(y') = (2\pi \sigma_Y^2)^{-1/2} \exp \left(-\frac {y'^2}{2 \sigma_Y^2} \right).
\end{equation}
The PDF of $X$ is more complicated because of its dependence on $Y$, but for any individual observation $y$ \eqref{eqn:prob_x} becomes a shifted Gaussian, giving:
\begin{align} 
    P_X(x|y) &= (2\pi \sigma_X^2 (1-\rho^2))^{-1/2} \exp \left( -\frac{ (x - \rho \tfrac{\sigma_X}{\sigma_Y} y)^2}{2 \sigma_X^2 (1-\rho^2)} \right),\text{ or} \nonumber \\
    P_X(y|y) &= (2\pi \sigma_X^2 (1-\rho^2))^{-1/2} \exp \left( -\frac{(1 - \rho \tfrac{\sigma_X}{\sigma_Y})^2 y^2}{2 \sigma_X^2 (1-\rho^2)} \right). \label{eqn:pxy}
\end{align}

\eqref{eqn:dkl1} then becomes:
\begin{align}
    \dkl(Y \Vert X) &= const(Y) - \int P_Y(y) \log(P_X(y)) \mathrm{d}y\ \nonumber \\
    &= const(Y) + \int (2\pi \sigma_Y^2)^{-1/2} \exp\left( -\frac{y^2}{2\sigma_Y^2} \right) \left( \log(2 \pi \sigma_X^2 (1-\rho^2))  + 
     \frac{(1 - \rho \tfrac{\sigma_X}{\sigma_Y})^2 y^2}{2 \sigma_X^2 (1-\rho^2)} \right) \mathrm{d}y \nonumber \\ 
    &= const(Y) + \log(\sigma_X^2 (1-\rho^2)) + \frac{(\sigma_Y - \rho \sigma_X)^2}{2 \sigma_X^2 (1-\rho^2)}. \label{eqn:kldiv_min}
\end{align}

Minimizing \eqref{eqn:kldiv_min} for $\sigma_X$ while holding $\sigma_Y$ and $\rho$ fixed is complicated, but solved numerically the optimal standard deviation ratio $\sigma_X \sigma_Y^{-1}$ is less than unity, reaching a minimum of about $0.66$ near $\rho = 0.4$ and increasing for both lower and higher values of correlation.  This is less intuitive than the $\sigma_X = \sigma_Y$ optimum of \eqref{eqn:amse}, but it still would smooth fine scales much less than the $\sigma_X = \sigma_Y \rho$ optimum of \eqref{eqn:mse1}.

Implementing \eqref{eqn:kldiv_min} as a loss function would be conceptually interesting, but this seems impractical because the expression has singular behaviour near $\rho=1$, where the implied random part of the prediction collapses to zero variance.

\section{Supplemental Verification}\label{app:verif}

\begin{figure}[t]
    \vskip 0.2in
    \begin{center}
    \centerline{\includegraphics[width=0.9\textwidth]{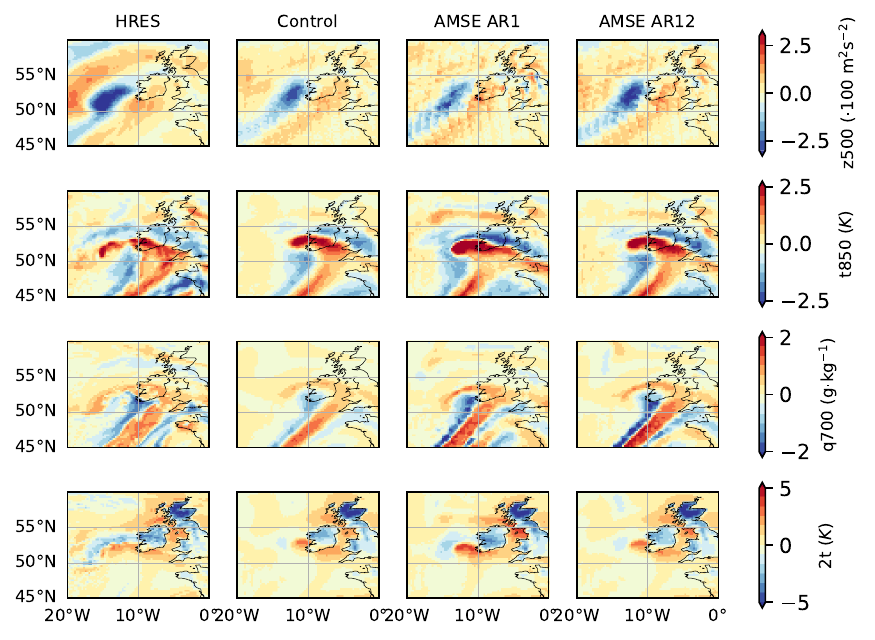}}
    \vskip -0.2in
    \caption{Visualization of high-pass filtered forecast and analysis fields for the forecast shown in \ref{fig:smooth_forecast}.}
    \label{fig:hpf}
    \end{center}
    \vskip -0.2in
\end{figure}

\subsection{Visualization} \label{app:hpf}

Figure \ref{fig:hpf} visualizes the high-wavenumber components of a sample forecast matching the winter storm Eunice prediction shown in figure \ref{fig:smooth_forecast}.  The applied filter fourth-order in spherical harmonic space, with the functional form:
\begin{equation}
\textrm{HPF}(k) = 1 - \frac{k_0^4}{k_0^4 + k^4},
\end{equation}
where $k$ is the total wavenumber and $k_0 = 50$ is the cutoff number, chosen to emphasize modes with length scales of 800 km and shorter.  Overall, the predictions of the control and AMSE-trained models show very similar structures, but training with \eqref{eqn:amse} as the loss function enhances the high-mode variability of the forecasts.

\subsection{Spectra} \label{app:spec}
\begin{figure}[t]
    \vskip 0.2in
    \begin{center}
    \centerline{\includegraphics[width=0.9\textwidth]{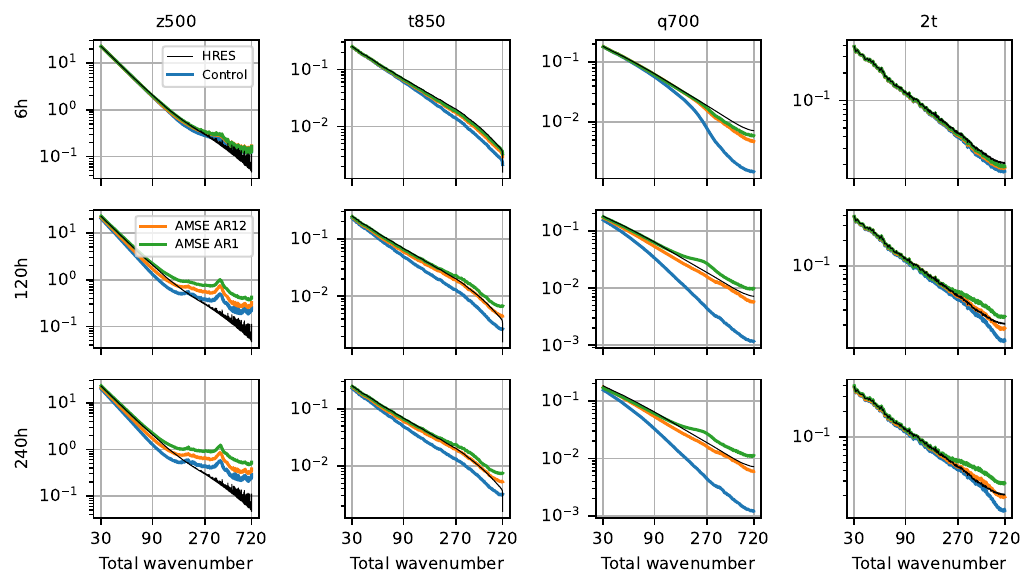}}
    \vskip -0.2in
    \caption{Amplitude spectral density for the variables of figure \ref{fig:ens_stats} at 6h, 120h, and 240h lead times.}
    \label{fig:spectra}
    \end{center}
    \vskip -0.2in
\end{figure}

Figure \ref{fig:spectra} shows the amplitude spectral density (square root of power spectral density, with units proportional to $1/\sqrt{cycle}$) at moderate to fine scales for several variables and lead times.  Because of the energy cascade in the atmosphere, the spectra of most variables follow power-law distributions.  Energy in the atmosphere is ultimately removed by turbulent, frictional dissipation, but no practical global atmospheric model can effectively resolve these scales.

Nonetheless, the available energy per total wavenumber varies over several orders of magnitude, and even large amplitude density differences between models can appear small on the typical log-log scales of these graphs.  The 2m temperature field shows very little smoothing compared to the analysis regardless of model because it is strongly affected by the local elevation, which is always supplied as a constant field.  

\subsection{Quantile/Quantile Plots} \label{app:qq}

\begin{figure}[htb]
    \vskip 0.2in
    \begin{center}
    \begin{tikzpicture}
        \node[inner sep=0pt] (qquv0) at (0,0)
            {\includegraphics[width=0.45\textwidth]{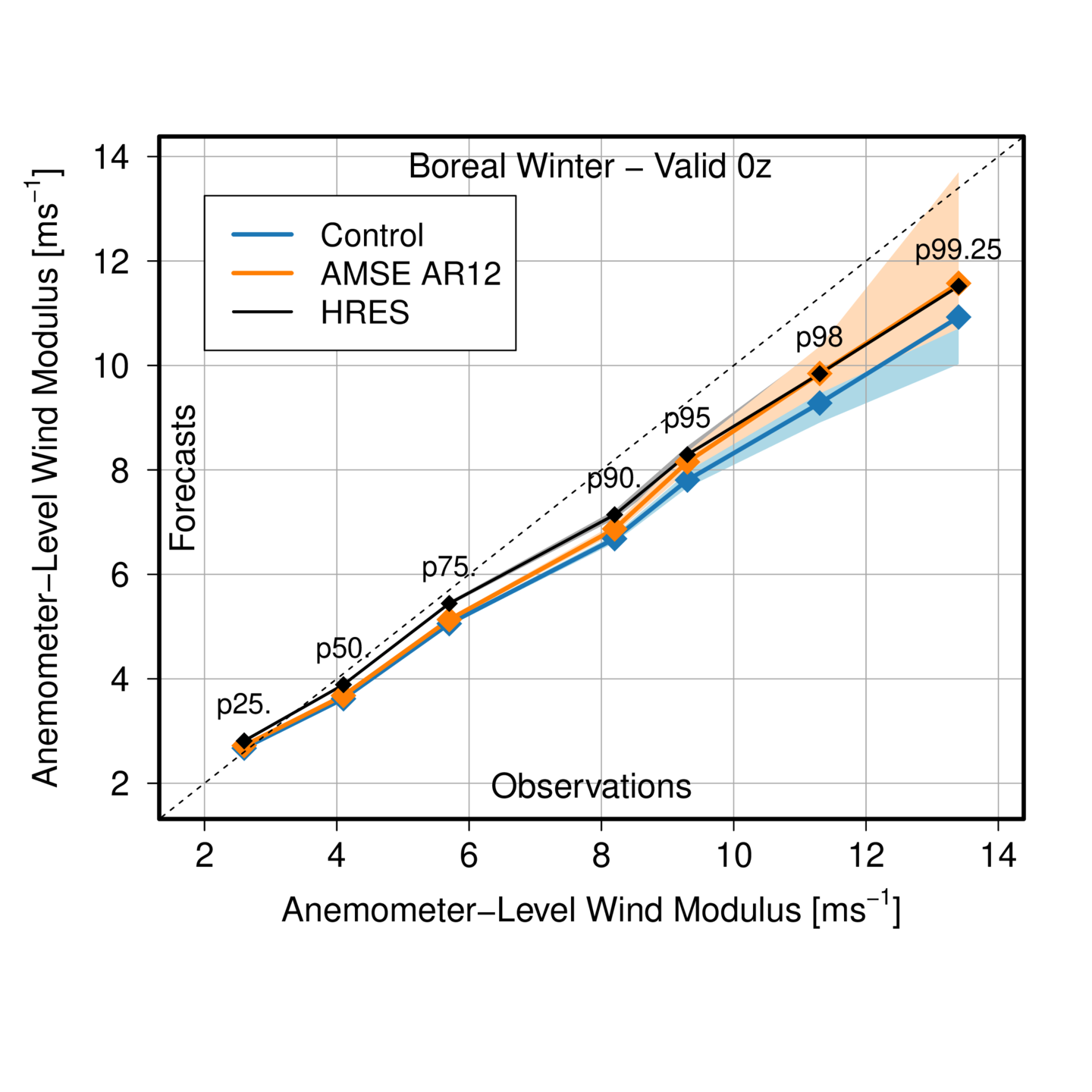}} ;
        \node[inner sep=0pt, right=0.0in of qquv0] (qquv1)  
            {\includegraphics[width=0.45\textwidth]{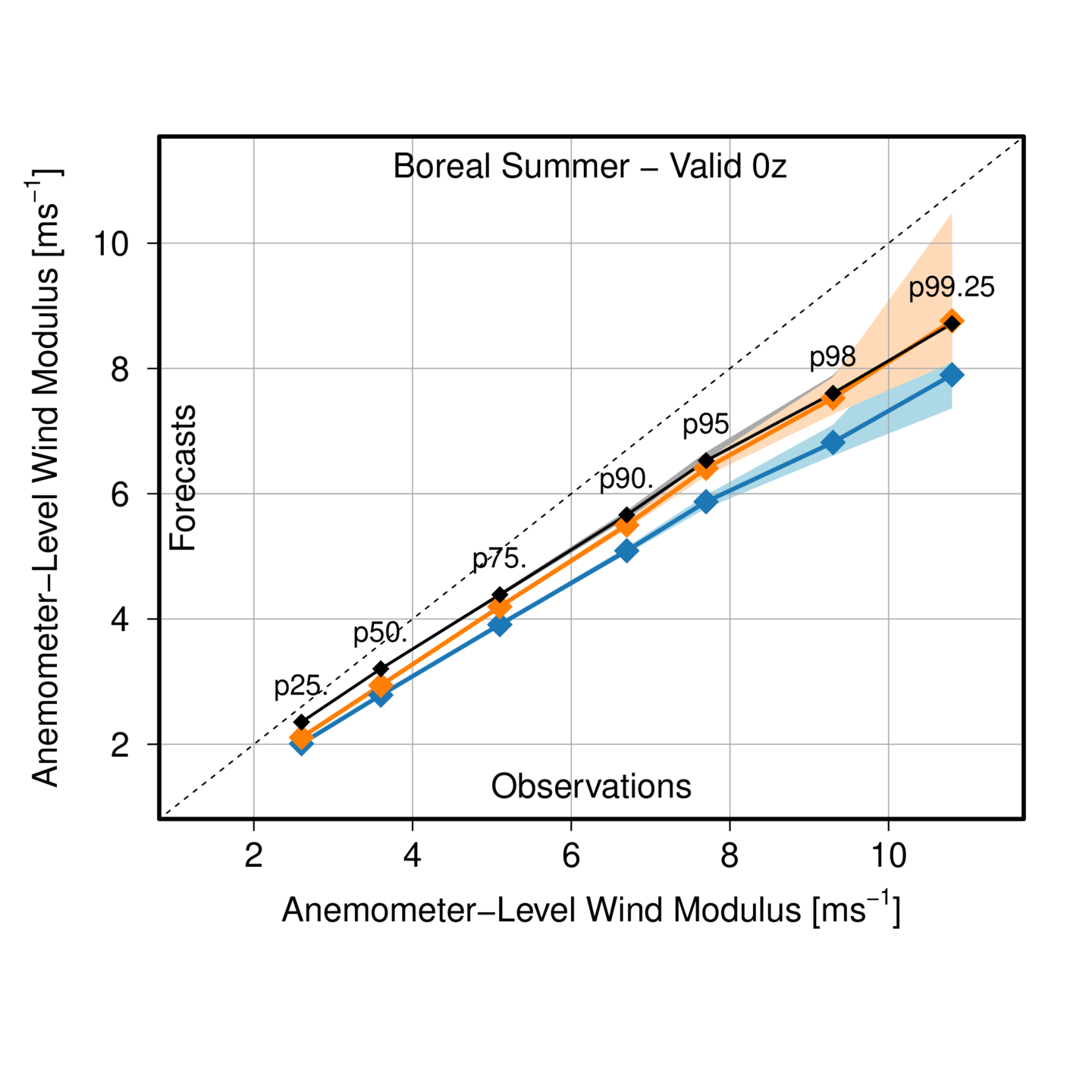}} ;
    \end{tikzpicture}
    \vskip -0.2in
    \caption{Quantile-quantile plots of 10 m wind speed at surface station locations for the North American domain.  At left, 1 Jan--30 March 2022 (boreal winter), and at right 20 June--19 September 2022 (boreal summer).  The control and AMSE AR12 points show model evaluations for 5-day forecasts.  The shaded region denotes confidence interval based on the Kolmogorov-Smirnov test.}
    \label{fig:qq_uv}
    \end{center}
    \vskip -0.2in
\end{figure}

\begin{figure}[htb]
    \vskip 0.2in
    \begin{center}
    \begin{tikzpicture}
        \node[inner sep=0pt] (qquv0) at (0,0)
            {\includegraphics[width=0.45\textwidth]{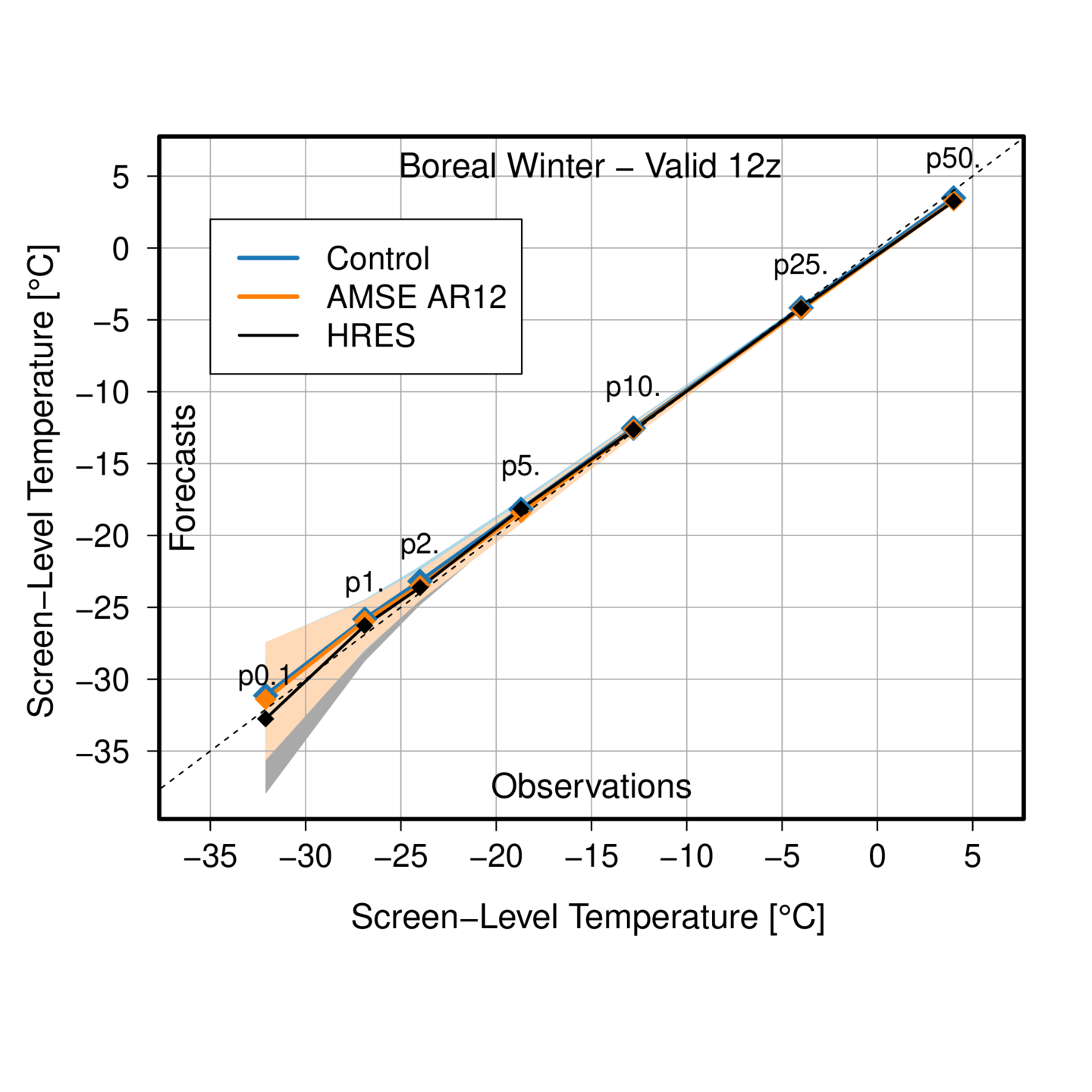}} ;
        \node[inner sep=0pt, right=0.0in of qquv0] (qquv1)  
            {\includegraphics[width=0.45\textwidth]{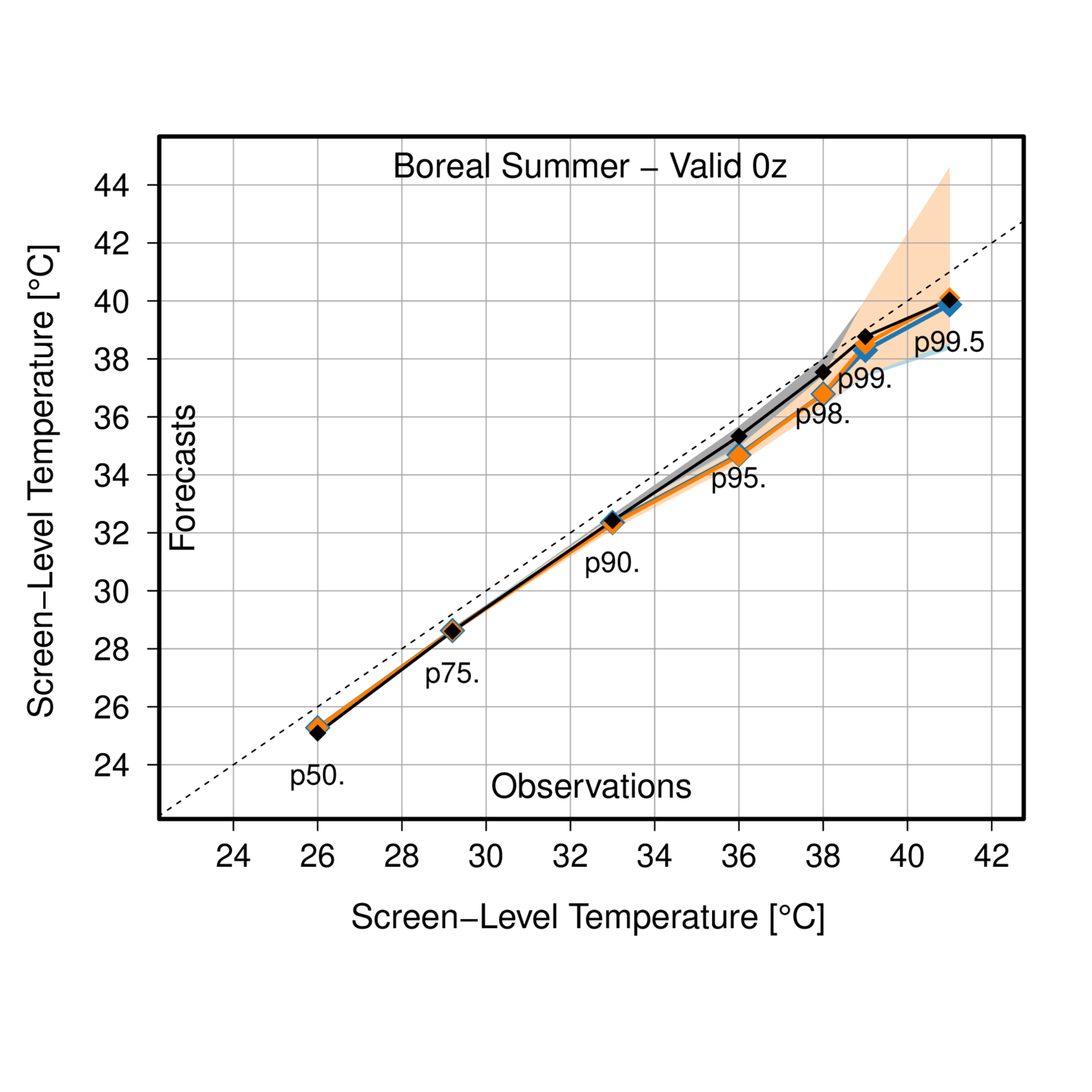}} ;
    \end{tikzpicture}
    \vskip -0.2in
    \caption{As in figure \ref{fig:qq_uv}, for 2m temperature.  Low percentiles (extreme cold) are shown for the Northern Hemisphere winter, and high percentiles (extreme heat) are shown for the Northern Hemisphere summer.}
    \label{fig:qq_tt}
    \end{center}
    \vskip -0.2in
\end{figure}

Quantile-quantile plots show a joint cumulative density function, and we use them here to evaluate the overall realism of the forecasts produced by the control and AMSE AR12 models independently of the forecast skill.  In figures \ref{fig:qq_uv} and \ref{fig:qq_tt}, the x-location of each point is the labelled percentile of North American weather station observations for Northern Hemisphere winter and summer periods.  The y-location of each point is the corresponding percentile for the HRES analysis or the 5-day forecasts produced by the control and AMSE AR12 models, interpolated to the station locations.  For example, in the left panel of figure \ref{fig:qq_uv}, the 98th percentile corresponds to an observed wind speed of about 11.5m/s, but the 98th percentile of the HRES analysis was about 10m/s.

The $y=x$ line on the quantile-quantile plot, shown as a dashed line in each panel, suggests that the forecast and observations have the same unconditional distributions when aggregated, and departures from the diagonal line indicate systematic underprediction or overprediction of extreme values.  In our case, figure \ref{fig:qq_uv} shows that the AMSE AR12 model has a more realistic representation of surface winds, matching the trends seen in the HRES data.  The control model produces noticeably weaker winds at all percentiles, showing a systematic shift in the distribution towards weaker surface winds, particularly in summer.

In contrast, figure \ref{fig:qq_tt} shows that the models are essentially equivalent in the distribution of 2m temperatures.  As discussed in section \ref{app:spec}, the 2m temperature field shows little smoothing in the control model, likely due to the strong influence of elevation on the surface temperature.  Improvements to the forecast of 2m temperature in the AMSE AR12 model are found more in the forecast skill (see figure \ref{fig:crps_scorecard}) than in the unconditional distribution of temperatures.

\subsection{Details of the Lagged Ensemble Verification}\label{app:ensemble}

\citet{lagged_ensemble} uses several metrics to evaluate the quality of the lagged ensembles.  In this work, we use the fair CRPS score, the ensemble root mean squared error (eRMSE), and the spread-error ratio (SER).  The eRMSE statistic is derived from its squared version (ensemble mean squared error), evaluated pointwise and integrated over the grid.  The SER statistic is the simple ratio of the integrated MSE and ensemble spread (unbiased estimate of variance), noting for emphasis that the ratio is taken after the grid-averaging.  For an ensemble of $N_e$ members ($x_{1\ldots N_e}$) evaluated over $N_{date}$ forecasts with verifying analysis $y$, the corresponding formulas are:
\begin{align} 
    \crps(x,y) &= \frac{1}{N_{date}} \sum_{d=1}^{N_{date}} \sum_{i,j} \mathrm{d}A(i,j)  \left(\frac{1}{N_e} \sum_{k=1}^{N_e} \vert x_k(i,j) - y(i,j) \vert + \right. \nonumber \\ 
    & \qquad \left. \frac{1}{2 N_e (N_e-1)} \sum_{k=1}^{N_e} \sum_{l=1}^{N_e} \vert x_k(i,j) - x_l(i,j) \vert\right), \label{eqn:crps} \\
    \erms(x,y) &= \left( \frac{1}{N_{date}} \sum_{d=1}^{N_{date}} \sum_{i,j} \mathrm{d}A(i,j) (\bar{x}(i,j) - y(i,j))^2 \right)^{1/2},\text{ and} \label{eqn:erms} \\
    \ser(x,y) &= \left( \frac{1}{N_{date}} \sum_{d=1}^{N_{date}} \frac{1}{N_e-1} \frac{\sum_{i,j} \mathrm{d}A(i,j) \sum_{k=1}^{N_e} (x_k(i,j) - \bar{x}(i,j))^2 }
                                                                      {\sum_{i,j} \mathrm{d}A(i,j) (\bar{x}(i,j) - y(i,j))^2} \right)^{1/2}, \label{eqn:ser}
\end{align}
where $\bar{x}(i,j) = N_e^{-1} \sum_k x_k(i,j)$ is the ensemble mean at the $(i,j)$ gridpoint.

\begin{figure}[htb]
    \vskip 0.2in
    \begin{center}
    \centerline{\includegraphics[width=0.9\textwidth]{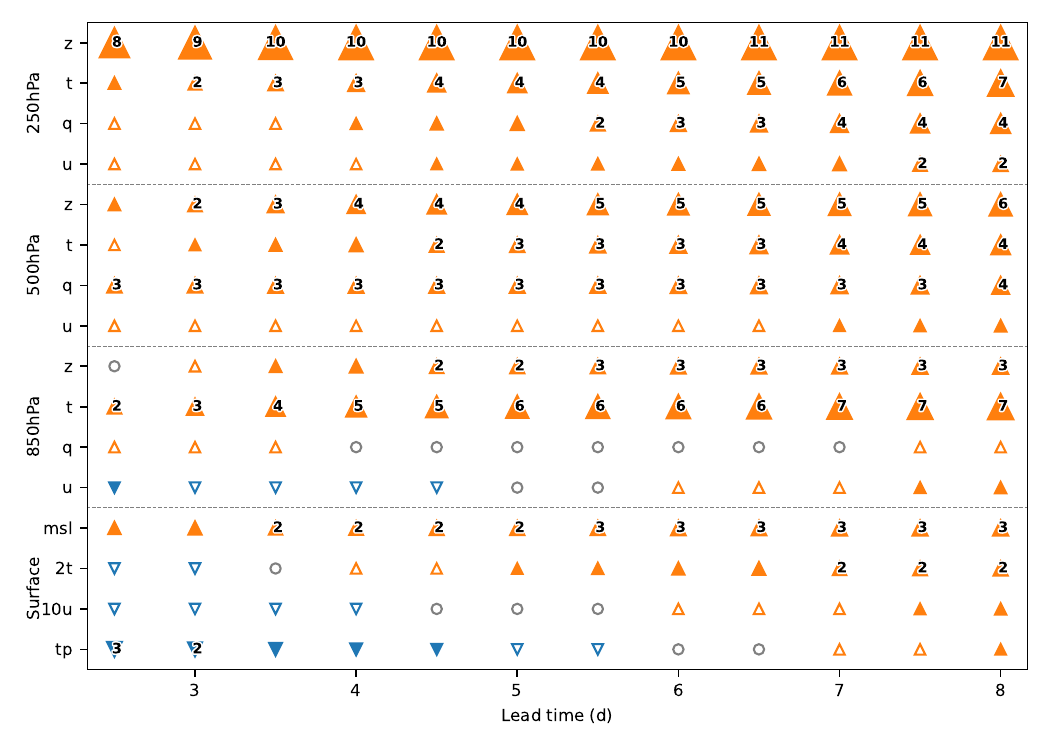}}
    \vskip -0.2in
    \caption{CRPS skill score (\% improvement), measured as the relative difference between the CRPS \eqref{eqn:crps} of the 12-step fine-tuned model and the CRPS of the control model, for a selection of variables and lead times.  Orange up-arrows show where the fine-tuned model performs better, blue down-arrows show where the control model performs better. Hollow arrows represent a difference of less than 1\%, and differences of 2\% or larger are marked.  Hollow circles mark values that are not statistically significant at the 90\% level.}
    \label{fig:crps_scorecard}
    \end{center}
    \vskip -0.2in
\end{figure}

\begin{figure}[htb]
    \vskip 0.2in
    \begin{center}
    \centerline{\includegraphics[width=0.9\textwidth]{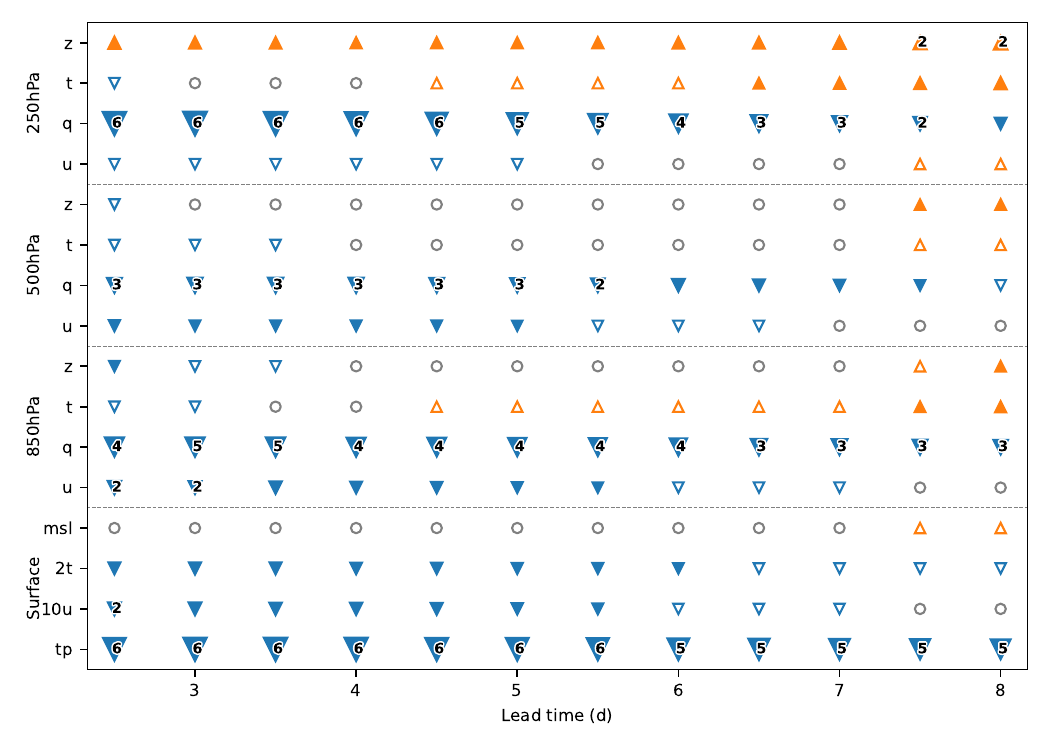}}
    \vskip -0.2in
    \caption{As figure \ref{fig:crps_scorecard}, for ensemble root mean squared error \eqref{eqn:erms}.}
    \label{fig:ermse_scorecard}
    \end{center}
    \vskip -0.2in
\end{figure}

Figures \ref{fig:crps_scorecard} and \ref{fig:ermse_scorecard} show the CRPS and eRMSE skill scores respectively of the lagged ensemble generated with AMSE AR12 compared to the lagged ensemble of the control model for the geopotential (z), temperature (t), specific humidity (q), and u-component of wind (u) at several elevations and for the mean sea level pressure (msl), 2-meter temperature (2t), u-component of 10m wind (10u), and 6h-accumulated precipitation (tp) at the surface. 

For these figures, statistical significance was determined by bootstrapping, sampling 1/3 of the total dates in each sample to give an average gap between dates of 36h.  The forecast skill of persistence (that is, the gain over a climatological forecast by predicting that everything will remain constant) decays very quickly over 36h, so samples so-spaced apart are reasonably independent of each other.

Overall, AMSE AR12 shows CRPS skill improvements for most variables and most lead times, but total precipitation shows only small improvements at long lead times and degradation at short lead times.  This is explained by the separation of modes in \eqref{eqn:smse_decomp} not being a natural one for precipitation.  Precipitation is often localized but always non-negative, and consequently its spectral decomposition does not really resemble the normally-distributed random values that give meaning to \eqref{eqn:smse_decomp} and \eqref{eqn:amse}.  

The eRMSE skill chart should be interpreted with caution.  The scores of \eqref{eqn:crps}--\eqref{eqn:ser} were developed for the case of an ideal ensemble, where members are statistically indistinguishable from each other and equally accurate in expectation.  This is not really the case for a lagged ensemble, where the shorter-duration members should be noticeably more accurate than longer-duration members and an ideal aggregation would separately weight each term.  This is not done by \citet{lagged_ensemble} for simplicity and to avoid free parameters, but we believe that the early lead-time smoothing in the control model makes equal-weighting more optimal for its lagged ensemble than for the lagged ensemble of AMSE AR12.  

For long lead times this advantage diminishes, where the relative degradation of forecast quality is much stronger between 0.5 days and 4.5 days than it is between 6 days and 10 days.  In this regime, AMSE AR12 begins to show eRMSE skill over the control ensemble.

\subsubsection{Unbiased Ensemble Root Mean Squared Error}

\begin{figure}[htb]
    \vskip 0.2in
    \begin{center}
    \centerline{\includegraphics[width=0.9\textwidth]{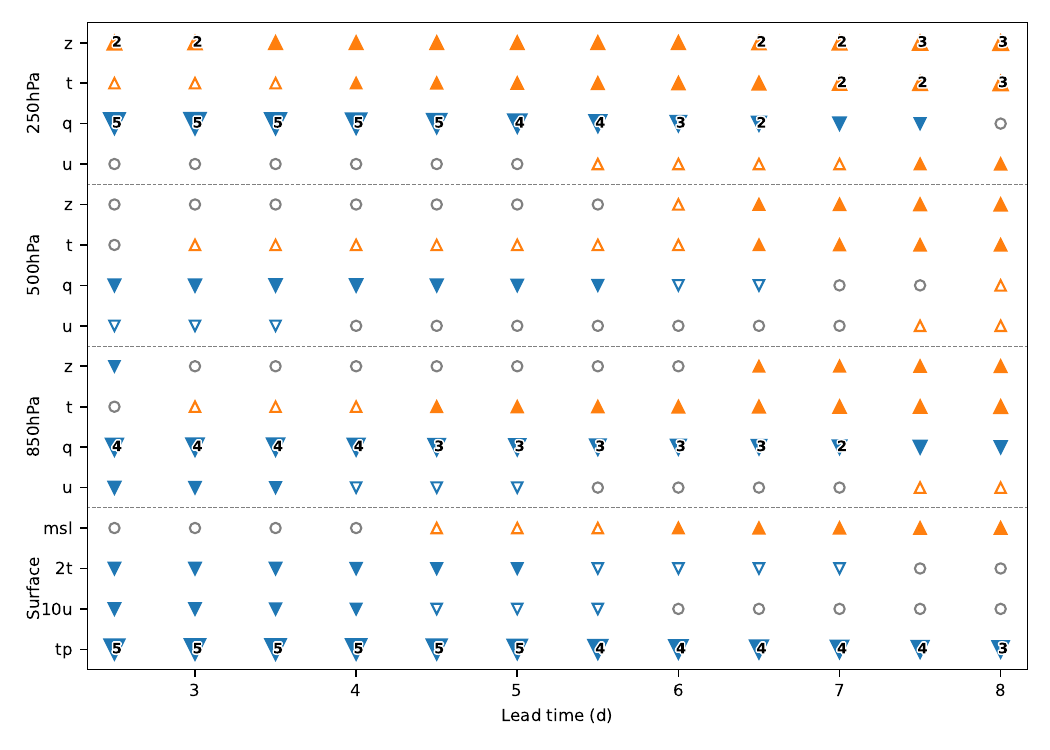}}
    \vskip -0.2in
    \caption{As figures \ref{fig:crps_scorecard} and \ref{fig:ermse_scorecard}, for the unbiased ensemble root mean squared error \eqref{eqn:ubrmse}.}
    \label{fig:ubermse_scorecard}
    \end{center}
    \vskip -0.2in
\end{figure}

The eRMSE formula of \eqref{eqn:erms} is a biased estimator of the true ensemble mean error, overestimating the error in proportion to the ensemble (sample) spread when the ensemble size is finite.  

Consider $N_e$ different realizations $({x_i})$ of a single variable drawn from $\normal(\mu,\sigma^2)$ when the ground-truth value is $0$.  Applying \eqref{eqn:erms} to this gives:
\begin{equation}
    \expect{(\mathrm{eMSE}(x,0))} = \expect \left( \left(\frac{1}{N_e} \sum_i x_i \right)^2 \right)  = \mu^2 + \frac{\sigma^2}{N_e}, \label{eqn:erms_bias}
\end{equation}
which overestimates the true ensemble mean squared error.  This overestimate is more severe for small ensembles such as the lagged ensemble configuration of section \ref{sec:ensemble}, where the ensemble size cannot be easily increased.

\citet{leutbecher_2008} proposes correcting this overestimate by subtracting the standard error term to give an unbiased estimator of the ensemble mean squared error with a finite sample size.  In the notation of \eqref{eqn:erms}, the corresponding root mean squared formula becomes:
\begin{align}
    \uberms(x,y) = \Bigg( \frac{1}{N_{date}} \sum_{d=1}^{N_{date}} & \sum_{i,j} \mathrm{d}A(i,j) \bigg(
                    \Big(\frac{1}{N_e} \sum_{k=1}^{N_e} x_k(i,j) - y(i,j) \Big)^2 - \nonumber \\
                           & \frac{1}{N_e(N_e-1)} \sum_{k=1}^{N_e} (x_k(i,j) - \bar{x}(i,j))^2
                    \bigg) \Bigg)^{1/2}, \label{eqn:ubrmse}
\end{align}
which performs this correction pointwise on the grid before computing the spatial average and taking the square root.

Implementing this adjustment slightly improves the scores of the AMSE-tuned model compared to the control model, and the corresponding ``scorecard'' is shown in figure \ref{fig:ubermse_scorecard}.

\clearpage

\subsection{Ablation Studies} \label{app:ablation}

\begin{figure}[htb]
    \vskip 0.2in
    \begin{center}
    \centerline{\includegraphics[width=0.9\textwidth]{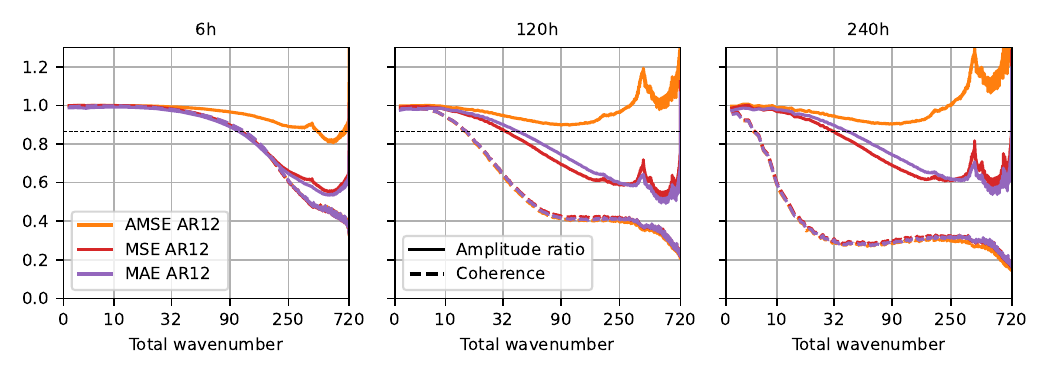}}
    \vskip -0.2in
    \caption{As figure \ref{fig:resolution}, for the comparison models of section \ref{app:ablation}.  Only the model trained with the AMSE error function retains sharpness to fine scales.}
    \label{fig:ablation_resolution}
    \end{center}
    \vskip -0.2in
\end{figure}

\begin{table}[tb]
    \caption{Cumulative ranked probability scores for the models fine-tuned in this study in the lagged ensemble configuration described in section \ref{sec:ensemble}, for the ``headline'' variables and levels in figure 1 of \citet{weatherbench2}.  Lower is better; the best score is bolded and the second-place score is italicized.} \label{tab:ablation}
    \vskip 0.10in
    \begin{center}
    \begin{small}
    \begin{tabular}{ccccccccccccc}
        \toprule
        Model & \multicolumn{3}{c}{z 500hPa ($\mathrm{m}^2 \mathrm{s}^{-2}$)} 
          & \multicolumn{3}{c}{t 850hPa ($\mathrm{K}$)}
          & \multicolumn{3}{c}{q 700hPa ($\mathrm{g}\cdot\mathrm{kg}^{-1}$)}
          & \multicolumn{3}{c}{u 850hPa ($\mathrm{m}\cdot\mathrm{s}^{-1}$)} \\
        & 2.5d & 5.0d & 7.5d 
          & 2.5d & 5.0d & 7.5d 
          & 2.5d & 5.0d & 7.5d 
          & 2.5d & 5.0d & 7.5d \\
        \midrule 
        Control & 31.038 & 84.315 & 162.481 
        & 0.428 & 0.691 & 1.003 
        & 0.357 & 0.523 & 0.652 
        & 0.823 & 1.340 & 1.904 \\ 
        MSE AR12 & 31.285 & 82.100 & 155.703 
        & 0.419 & 0.664 & 0.949 
        & 0.356 & 0.526 & 0.652 
        & \textit{0.819} & \textit{1.335} & 1.886 \\
        MAE AR12 & \textbf{29.969} & \textit{80.621} & \textit{155.361} 
        & \textbf{0.410} & \textit{0.654} & \textit{0.947} 
        & \textbf{0.340} & \textbf{0.499} & \textbf{0.624} 
        & \textbf{0.811} & \textbf{1.313} & \textbf{1.859} \\
        AMSE AR1 & 33.720 & 94.703 & 186.202 
        & 0.422 & 0.721 & 1.078 
        & 0.354 & 0.558 & 0.721 
        & 0.863 & 1.485 & 2.115 \\
        AMSE AR12& \textit{30.565} & \textbf{80.469} & \textbf{153.267}
        & \textit{0.418} & \textbf{0.653} & \textbf{0.935} 
        & \textit{0.347} & \textit{0.510} & \textit{0.634} 
        & 0.832 & 1.341 & \textit{1.882} \\
        \bottomrule
    \end{tabular}
    \end{small}
    \end{center}
    \vskip -0.15in
\end{table}

To ensure that the results of this study are not simply an artifact of increasing the model's overall training time, we compare against two additional fine-tunings:
\begin{enumerate}
    \item MSE AR12 implements the fine-tuning schedule of table \ref{tab:finetune} with the unmodified mean squared error loss function, as with GraphCast's principal training.
    \item MAE AR12 implements the fine-tuning schedule with a mean absolute error loss function, preserving the per-variable and per-level weightings of error.
\end{enumerate}

Figure \ref{fig:ablation_resolution} shows the aggregated per-wavenumber performance of these models, and table \ref{tab:ablation} evaluates their CRPS for a selection of variables, levels, and lead times in the lagged ensemble configuration.

Both models still show excessive smoothing of fine scales, but training with mean absolute error moderately improves sharpness at the medium scales (wavenumbers 32--200 for longer lead times, corresponding to length scales of 1250--250 kilometers).  

The excessive smoothing of the MSE-trained model is expected from section \ref{subsec:optimal}, but that argument does not directly apply to the mean absolute error loss function.  However, we can still understand this behaviour intuitively.  A model that is optimal under the mean absolute error predicts the mean of a distribution, and at longer lead times fine scales are less predictable than coarser scales.  Therefore, the prediction of the median future should be smoother than its realization.

Even the moderate improvement to sharpness for the MAE-trained model results in improvements to the CRPS of the lagged ensemble, as shown in table \ref{tab:ablation}.

%%%%%%%%%%%%%%%%%%%%%%%%%%%%%%%%%%%%%%%%%%%%%%%%%%%%%%%%%%%%
%%%%%%%%%%%%%%%%%%%%%%%%%%%%%%%%%%%%%%%%%%%%%%%%%%%%%%%%%%%%%%%%%%%%%%%%%%%%%%%

\end{document}